\documentclass[12pt]{article}
\usepackage[utf8]{inputenc}
\usepackage[T1]{fontenc}
\usepackage{lmodern}
\usepackage{float}
\usepackage{csquotes}
\usepackage{svg}
\usepackage[ruled,vlined]{algorithm2e}
\usepackage{amsmath,amssymb,amsthm}
\usepackage{graphicx}
\usepackage{multirow}
\usepackage{booktabs}
\usepackage{pdfpages}
\usepackage{caption}
\usepackage{subcaption}
\usepackage[colorlinks=true,linkcolor=blue,citecolor=blue,urlcolor=blue]{hyperref}\usepackage{enumitem}
\usepackage{cleveref}
\usepackage{geometry}
\usepackage{microtype}

\usepackage[table]{xcolor} 
\definecolor{cstripe}{HTML}{EAF2F5}

\usepackage[style=apa]{biblatex}
\usepackage{lineno}

\usepackage[american]{babel}

\DeclareLanguageMapping{american}{american-apa}

\DeclareCiteCommand{\cite}
  {\usebibmacro{prenote}}
  {\usebibmacro{citeindex}%
   \printtext[bibhyperref]{\usebibmacro{cite}}}
  {\multicitedelim}
  {\usebibmacro{postnote}}

\DeclareCiteCommand*{\cite}
  {\usebibmacro{prenote}}
  {\usebibmacro{citeindex}%
   \printtext[bibhyperref]{\usebibmacro{citeyear}}}
  {\multicitedelim}
  {\usebibmacro{postnote}}

\DeclareCiteCommand{\parencite}[\mkbibparens]
  {\usebibmacro{prenote}}
  {\usebibmacro{citeindex}%
    \printtext[bibhyperref]{\usebibmacro{cite}}}
  {\multicitedelim}
  {\usebibmacro{postnote}}

\DeclareCiteCommand*{\parencite}[\mkbibparens]
  {\usebibmacro{prenote}}
  {\usebibmacro{citeindex}%
    \printtext[bibhyperref]{\usebibmacro{citeyear}}}
  {\multicitedelim}
  {\usebibmacro{postnote}}

\DeclareCiteCommand{\footcite}[\mkbibfootnote]
  {\usebibmacro{prenote}}
  {\usebibmacro{citeindex}%
  \printtext[bibhyperref]{ \usebibmacro{cite}}}
  {\multicitedelim}
  {\usebibmacro{postnote}}

\DeclareCiteCommand{\footcitetext}[\mkbibfootnotetext]
  {\usebibmacro{prenote}}
  {\usebibmacro{citeindex}%
   \printtext[bibhyperref]{\usebibmacro{cite}}}
  {\multicitedelim}
  {\usebibmacro{postnote}}

\DeclareCiteCommand{\textcite}
  {\boolfalse{cbx:parens}}
  {\usebibmacro{citeindex}%
   \printtext[bibhyperref]{\usebibmacro{textcite}}}
  {\ifbool{cbx:parens}
     {\bibcloseparen\global\boolfalse{cbx:parens}}
     {}%
   \multicitedelim}
  {\usebibmacro{textcite:postnote}}

\title{}
\author{}
\date{}

\begin{document}
\vspace{-3cm}
\begin{center}
{\LARGE\bfseries
How to Optimize Multispecies Set Predictions in Presence–Absence Modeling ? \par
} 
\vspace{0.5cm}
Sébastien Gigot--Léandri$\color{blue} ^{1,2}$ | Gaétan Morand$\color{blue} ^{2}$ | Alexis Joly$\color{blue} ^{1}$ | François Munoz$\color{blue} ^4$ \\
David Mouillot$\color{blue} ^{2}$ | Christophe Botella$\color{blue} ^{1}$ | Maximilien Servajean$\color{blue} ^{1,3}$     
\end{center}

\vspace{0.5cm}
\noindent
{\small
$\color{blue} ^1$UMR LIRMM, University of Montpellier, Inria, CNRS, France\\
$\color{blue} ^2$UMR Marbec, University of Montpellier, IRD, CNRS, Ifremer\\
$\color{blue} ^3$University of Montpellier Paul Valéry, Montpellier, France\\
$\color{blue} ^4$Laboratoire de Biométrie et de Biologie Evolutive, Université Lyon 1, Villeurbanne, France 
}
\vspace{1cm}

\subsubsection*{Author Contributions}

{\footnotesize
SGL conceived and designed the study, developed the MaxExp optimization framework, implemented all core algorithms, and led the analyses. GM developed and implemented the modeling pipeline for the Reef Life Survey case study. AJ, MS, CB, and FM contributed to the study design, methodological refinements, and interpretation. FM and DM provided ecological expertise and guidance on the ecological relevance of model evaluation and species assemblage predictions. SGL wrote the first draft of the manuscript, and all authors contributed to manuscript revision.
}
\subsubsection*{Data Availability Statement}
{\footnotesize
All results and graphs in this study are reproductible via the available code and the ready to download model outputs and ground truth data : \\ 
\textbf{Code :} Github: \url{https://github.com/sebastiengl/score_maximisation} \\
\textbf{Model outputs / Ground truth:} Seafile : \url{https://lab.plantnet.org/seafile/d/822f01a64a614759b11b/}\\ \\
The related studies and data supporting the findings of this paper are also openly available : \\
\textbf{Case Study 1 :} Paper: \cite{Picek2024} | Challenge \& Model: \url{https://www.kaggle.com/code/picekl/sentinel-landsat-bioclim-baseline-0-31626}\\
\textbf{Case Study 2 :} Paper: \cite{Morand2025} | RLS: \cite{Edgar2014}\\
\textbf{Case Study 3 :} Paper: \cite{Zipkin2023a} | BBS: \cite{Hudson2017} | eBird: \cite{Sullivan2014} | Modified Model : \url{https://github.com/sebastiengl/CS3_modified_model}
}
\vfill 
\begin{flushright}
\subsubsection*{Keywords}
{\footnotesize
species distribution models, presence-absence predictions, binarization methods
}
\end{flushright}
\vspace{0.5cm}
\begin{minipage}{0.5\textwidth}
\scriptsize
\textbf{Number of words in Abstract:} 146\\
\textbf{Number of words in Main Text:} 4165\\
\textbf{Number of References:} 37\\
\textbf{Number of Figures \& Tables:} 6
\end{minipage}
\begin{minipage}{0.5\textwidth}
\raggedleft\scriptsize
\textbf{Sébastien Gigot--Léeandri}\\
\textbf{E-mail:} s.gigotleandri@lirmm.fr\\
\textbf{Tel:} +33 6 75 00 89 93
\end{minipage}

\newpage

\begin{figure}[H]
    \centering
    \includegraphics[width=\linewidth]{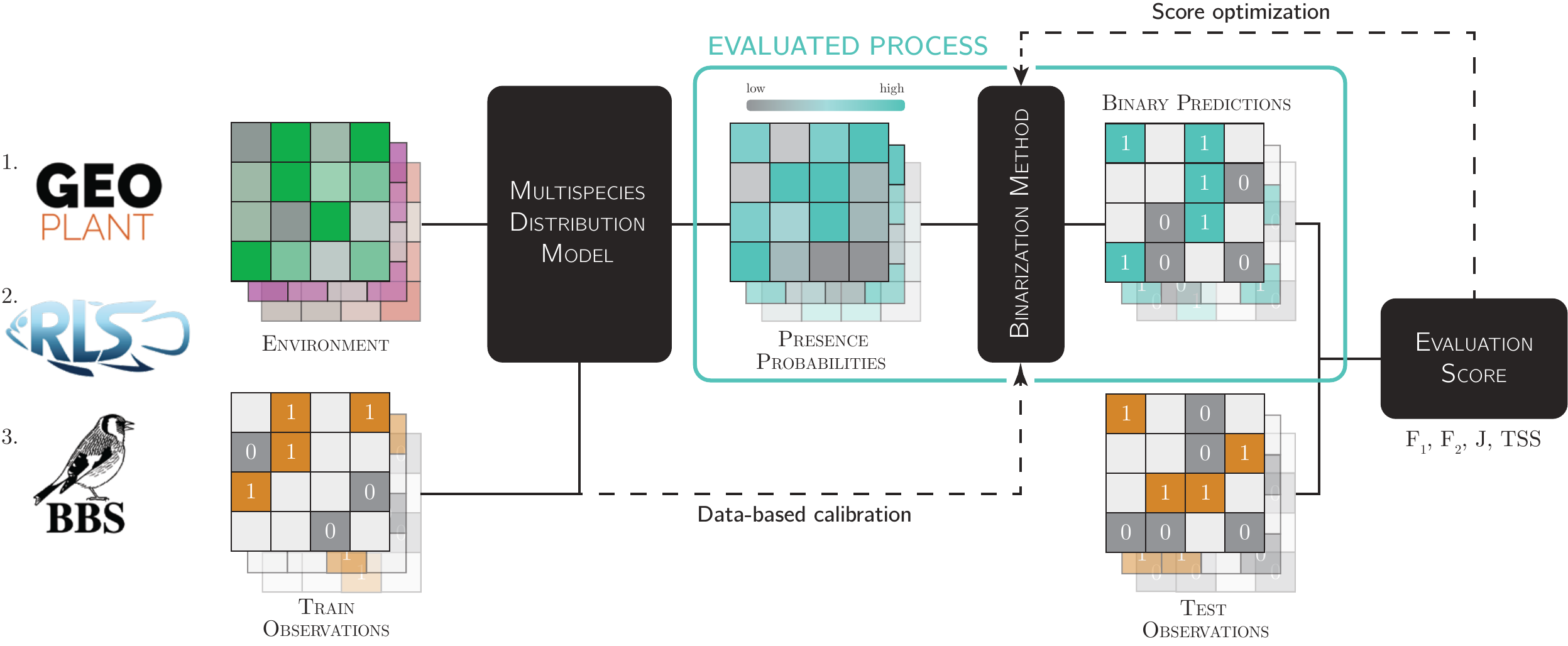}
    \captionsetup{justification=centering}
    \caption*{Across three case studies, we evaluate newly developed unsupervised binarization methods against
established calibration-based approaches. MaxExp, the framework introduced here, combines score
maximization with the advantages of unsupervised optimization and delivers improved multispecies
predictions across ecosystems.}
    \label{graphical_abstract}
\end{figure}

\section{Abstract}

Species distribution models (SDMs) commonly produce probabilistic occurrence predictions that must be converted into binary presence–absence maps for ecological inference and conservation planning. However, this binarization step is typically heuristic and can substantially distort estimates of species prevalence and community composition. We present MaxExp, a decision-driven binarization framework that selects the most probable species assemblage by directly maximizing a chosen evaluation metric. MaxExp requires no calibration data and is flexible across several scores. We also introduce the Set Size Expectation (SSE) method, a computationally efficient alternative that predicts assemblages based on expected species richness. Using three case studies spanning diverse taxa, species counts, and performance metrics, we show that MaxExp consistently matches or surpasses widely used thresholding and calibration methods, especially under strong class imbalance and high rarity. SSE offers a simpler yet competitive option. Together, these methods provide robust, reproducible tools for multispecies SDM binarization.

\newpage
\section{Introduction}
\begin{figure}[H]
    \centering
    \includegraphics[width=0.9\linewidth]{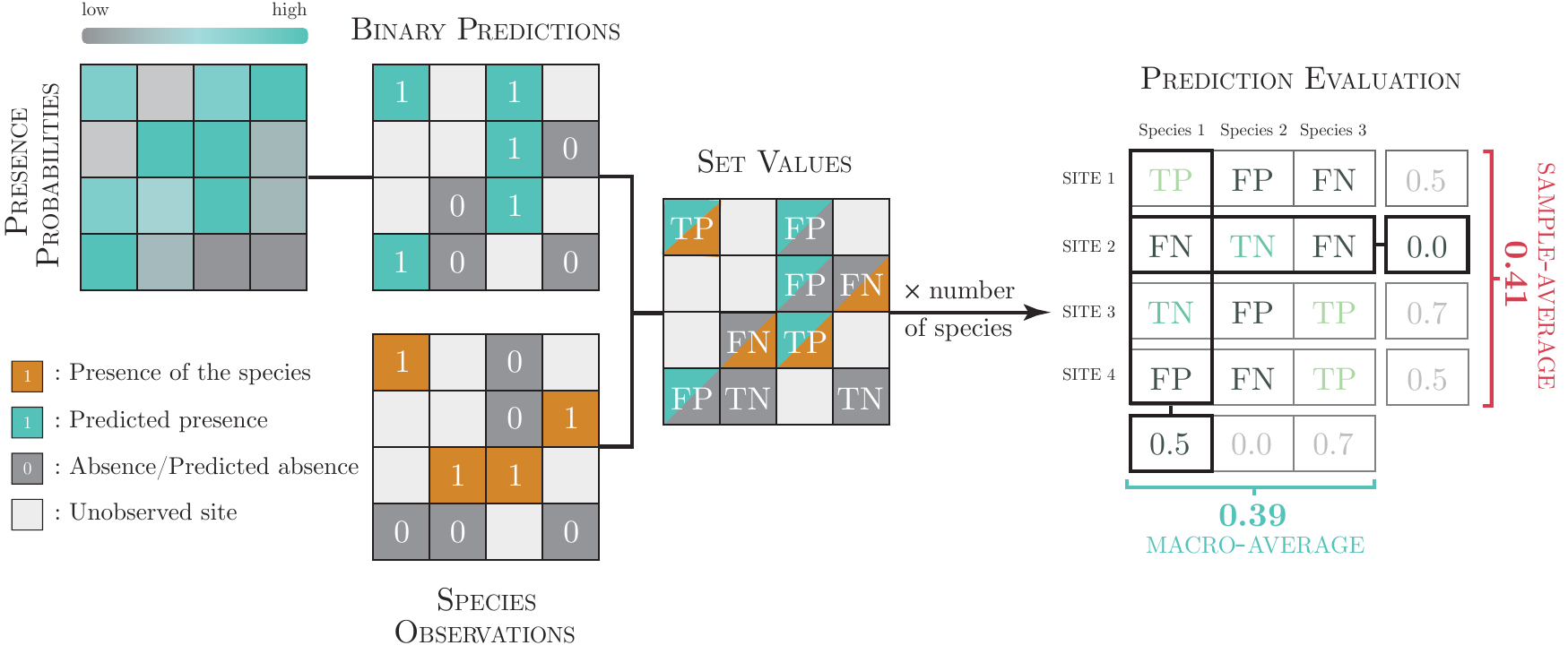}
    \caption{Evaluation pipeline of species presence/absence predictions. The spatial probability estimates (in blue shade) outputs by the model for a given species are converted into binary presence/absence predictions (blue/grey squares) where observations have been made. The results are then compared to the observed occurrence data (orange for presence, grey for absence). From this comparison, we get set values for each species at each site considered  (TP: True Positive, TN: True Negative, FP: False Positive and FN: False Negative). The prediction score can be then calculated in 2 main ways: macro-averaging, i.e., the mean of scores calculated individually for each species (blue), or the sample-averaging, calculating the mean of species scores computed at each site (red).}
    \label{set_values}
\end{figure}

Biodiversity is declining globally, so urgent conservation and management strategies require a deeper understanding of the intricate species dynamics across space an time (\cite{Dirzo2014}). In this context, Species Distribution Models (SDMs) are crucial tools for biodiversity monitoring and ecological assessment, as they can help understand species ecological niches, and predict species habitat suitability (\cite{Guisan2013}, \cite{Guisan2005}). \\
Modern SDMs, like JSDM (\cite{Wilkinson2021}), DeepSDM (\cite{Deneu2021}) or site occupancy models (\cite{Chambert2015}) address complex multispecies prediction tasks depending on diverse covariates, producing either suitability indices or, when calibrated with presence–absence data, presence probability estimates. Many practical ecological applications however still require binary decisions—namely, determining whether a species (or a set of species) should be present or absent at a given site (\cite{Araujo2006}, \cite{Luoto2006}, \cite{Jimenez2007}). Such decisions are critical for achieving various objectives, including estimating species ranges (\cite{Hellegers2025}), predicting species assemblages (\cite{Guisan2011}), mapping potential distributions of protected species (\cite{Zizka2020}), or forecasting species richness under scenarios of environmental changes (\cite{Guisan2011}). Thus, even with accurate presence probability estimates, binarizing the model outputs is necessary.
Many widely used evaluation scores in ecology—such as $F_1$-score, the Jaccard index, or the True Skill Statistic (TSS)—are based on binary predictions and set values (i.e., the number of True Positives, True Negatives, False Positives, False Negatives). These scores evaluate the agreement between predicted and observed presence/absence patterns, thereby assessing the combined efficiency of prediction and binarization steps (see \Cref{set_values}). However, this step introduces uncertainty and biases due to the information loss and arbitrary decisions inherent to thresholding (\cite{Nenzen2011}).\\\\

\begin{figure}[H]
    \centering
    \begin{subfigure}[t]{0.45\textwidth}
        \centering
        \includegraphics[width=\linewidth, keepaspectratio]{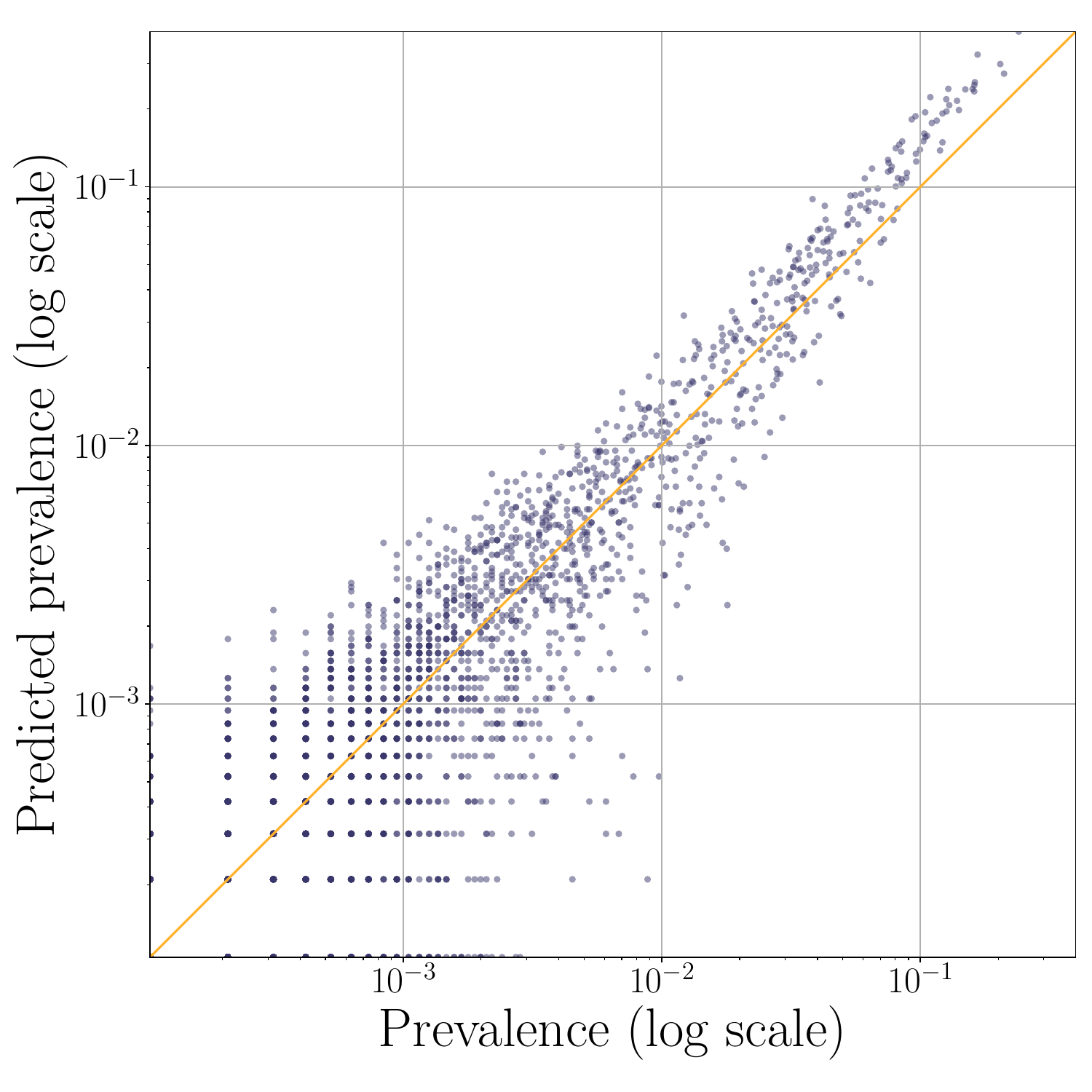}
        \caption{Species prevalences of the sample-average $F_1$-optimized predictions (y-axis) compared to true species prevalence (x-axis). Log-scale is used to underline how prevalence error has much higher variance for rare species.}
        \label{F1_prev}
    \end{subfigure}
    \hspace{0.5cm}
    \begin{subfigure}[t]{0.45\textwidth}
        \centering
        \includegraphics[width=\linewidth, keepaspectratio]{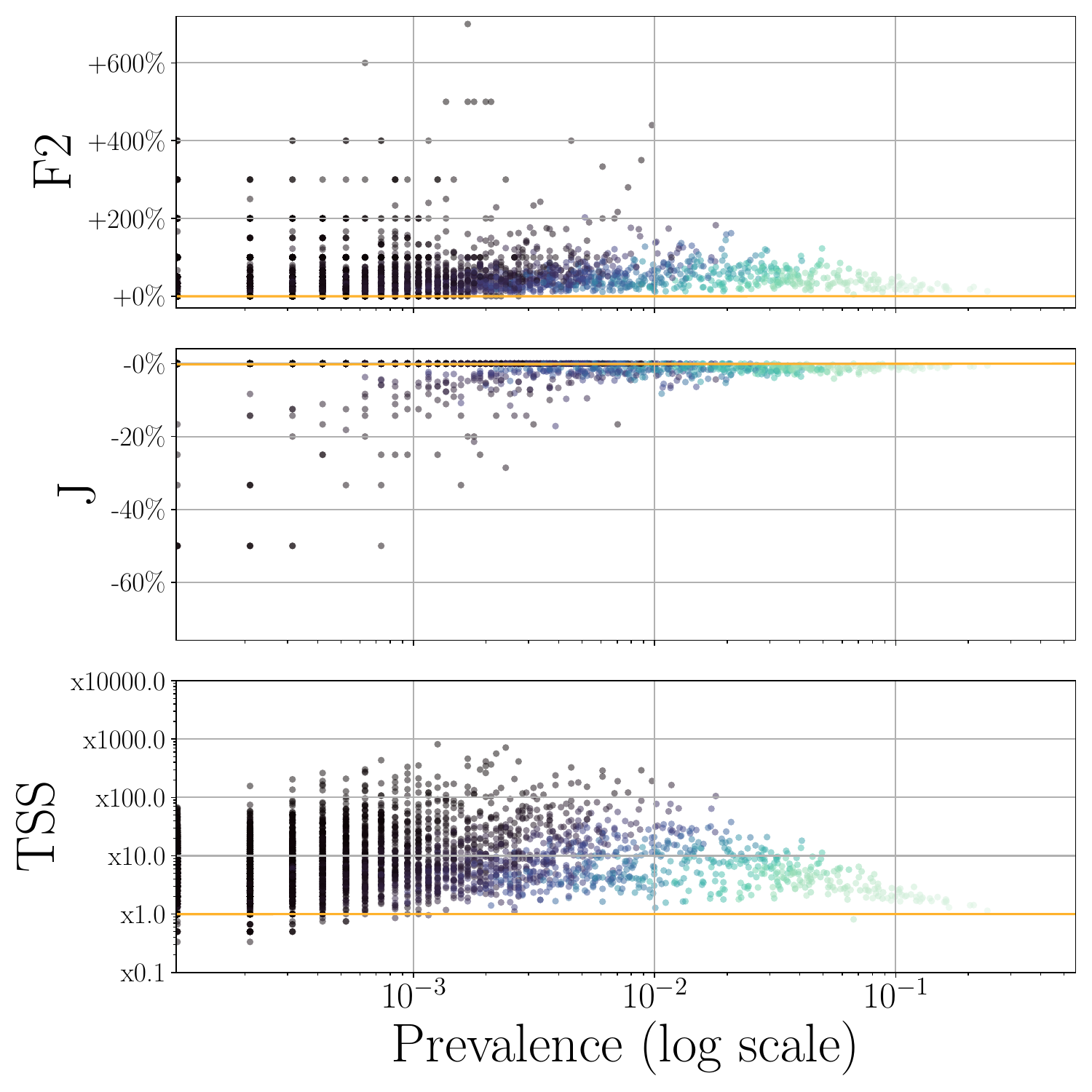}
        \caption{Change in predicted prevalences for different scores. To reduce displayed impact from prediction variance error, the results are this time compared directly to the prevalence predicted with $F_1$-score via their ratio (y-axis), in function of true prevalence (x-axis). The color gradient follows the predicted prevalence under $F_1$-score.}
        \label{ratio_prev}
    \end{subfigure}
    \caption{Study of the predicted species prevalence maximizing sample-average scores for the GeoPlant 2024 data (See Case 1 for more details of the model and dataset). Log values of predicted and true number of occupied sites + 1 are shown. The choice of evaluation score markedly influences predictive outcomes: the $F_1$-score produces balanced results with strong correlation to prevalence ($R_2: 0.69$, $R_2: 0.82$ in log-log scale). The $F_2$-score over-predicts the number of occupied sites, particularly for rare species. The Jaccard index yields the opposite effect with smaller magnitude, with rare species predicted in $\sim$ 20\% less sites. The TSS generates a more complex over-prediction pattern, with a curve-shaped ratio centered around a factor 10.}
    \label{metrics_prev}
\end{figure}

In practice, one must select one or several evaluation scores, and this choice has important consequences for the resulting species predictions of a selected SDM. Each score is sensitive to different types of error: For instance, some are more lenient towards false positives, while others are more conservative on occurrence prediction. As such, maximizing different scores can lead to broadly different predictions. For instance, \Cref{metrics_prev} depicts the influence of score choice on predicted prevalence (the proportion of sites where the species is predicted present). Therefore a score may be chosen to meet a specific study objective, such as predicting the species composition at a given site while minimizing the chance of missing rare species, or constructing a unbiased map of species richness. In the same manner, binary decisions must then be taken in accordance with that objective.

Therefore, identifying a general binarization method that performs well across a diverse range of scores remains both a critical and major challenge. Although the thresholding problem has often been addressed in the literature, most of the studies considered approaches relying on additional observation data (\cite{Petitpierre2017}, \cite{Li2013}, \cite{Zurell2020}), which entail increased risks of overfitting calibration (\cite{CholletRamampiandra2023}, \cite{Randin2006}), and substantial computational burden — particularly as the number of species increases. For rare species, allocating a significant portion of the already limited presence data to threshold calibration can further increase prediction uncertainties. (\cite{Nenzen2011}) noted that the SDM binarization step is one of the least explored source of error for these models whereas there are potentially many applicable rules for this binarization, and, as a consequence, the binarization method is often chosen arbitrarily. \\

The present study introduces a generic framework that optimizes the binarization process for multiple species by maximizing the expected of the provided evaluation score. In contrast to supervised thresholding methods, our method does not require more data than the data used to train the model, is applicable to a wide range of SDM evaluation scores, and achieves equivalent or superior performance compared to state-of-the-art methods across three distinct case studies and modeling approaches evaluated with 4 different sample-average scores. This demonstrates both the robustness and practical relevance of our new method for a myriad of ecological applications where binary presence–absence model outputs underpin conservation and management decisions.

\section{Materials \& Methods}
This study addresses the optimization of presence-absence multispecies predictions using sample-averaged (or horizontal) scores—that is, scores calculated for each site across all species and then averaged over all sites to produce a global performance measure. This approach contrasts with the more conventional use of macro-averaged (or vertical) scores, where scores are computed separately for each species and then averaged across species (\Cref{set_values}). In both frameworks, the optimization targets the agreement between predicted and observed species compositions (e.g. through $F_1$-score). The horizontal perspective (sample-average), however, emphasizes the accuracy of predicted species assemblages at site level, whereas the vertical perspective (macro-average) emphasizes the accuracy of species prevalence across the sites. By focusing on per-site assemblages and sample-averaged scores, our approach addresses an underexplored area in the literature with the potential to provide more accurate and realistic species assemblage predictions. Noticeably, the introduced framework is transferable to optimize SDM-binarization based on vertical scores as well.

\subsection{Problem formulation}

In the context of multispecies distribution models, a prediction is made at each site, and a local score can be calculated using a given evaluation score. The final score is then computed as the mean of local scores across sites.
We define the problem of species assemblage prediction in a given sampled site as a solvable optimization problem. This sample site is linked to an input environment $x \in \mathcal{X}$. Let us denote $y \in \mathcal{Y}$ the species assemblage associated to this input environment $x$, with $\mathcal{Y}=\mathcal{P}(\{ 1, \dots, N\})$ being the set of all possible species assemblages - i.e. the set of subsets of ${1, ..., N}$. The assemblage $y$ is supposed to be sampled from a probability measure $\mathbb{P}_{X =x}(Y)$.

Now, let \(U:\mathcal{Y}\times\mathcal{Y}\to\mathbb{R}\)  be a similarity function measuring how close two assemblages are (e.g. F1-score, Jaccard or TSS). This function must satisfy the following assumption:
\begin{itemize}
    \item[\textbf{A1}]: $U$ only depends on the number of true positives (TP), false positives (FP), true negatives (TN) and false negatives (FN). It increases with TP and TN, but decreases with FN and FP. This assumption is verified by most of the popular metrics in the literature. \\
\end{itemize}
We will also consider a second assumption as:
\begin{itemize}
    \item[\textbf{A2}]: the presences of all species are considered independent events from each other.
\end{itemize}

Our goal is to find the optimal species set $y^\star$ maximizing the expected agreement between the prediction $y'$ and the species assemblage $y$ in regard to $U$, which is defined as:

\begin{equation}
y^\star = \arg\max_{y'\in\mathcal{Y}} \mathbb{E}_{Y\mid X=x}[ U(Y,y') ]
= \arg\max_{y'\in\mathcal{Y}} \sum_{y\in\mathcal{Y}} \mathbb{P}_{X=x}(Y=y)\, U(y',y).
\end{equation}

The core challenge stems from the cardinality of $\mathcal{Y}$, which grows exponentially with the number of species $N$. Since $\mathcal{Y}$ is iterated over twice (for both $y'$ and $y$, representing the predicted and true species sets, respectively), the problem becomes computationally intractable with a total complexity of $\mathcal{O}(2^N)$. The objective of the present framework is to address this issue under some assumptions to render the problem computationally solvable. A similar strategy was used in (\cite{Nan2012}), but their approach was limited to $F_\beta$ measures and was not designed in a context of ecological application. \\

Solving the optimization problem under the additional assumptions that the prediction returns $k$ species and that species are independent under \textbf{A2} becomes trivial. To see this, consider an assemblage of size $n$ species. We have the following relations :

\begin{equation}
 \mathrm{FP} = k - \mathrm{TP} \quad \quad
 \mathrm{FN} = n - \mathrm{TP} \quad \quad
 \mathrm{TN} = N - k - n + \mathrm{TP}
 \end{equation}

In other words, all quantities depend on TP. Hence, considering assumptions \textbf{A1} and \textbf{A2} together, maximizing $U$ leads to the selection of the $k$ most probable species. Since this strategy is optimal for any size $n$, it is globally optimal. Had the species not been independent, the $k$ most likely species could potentially depend on the set size $n$. Hence, the expected score $\mathbb{E}_{Y|X=x}[U(Y,.)]$ is maximized by the topk predictor. The overall problem can therefore be rephrased as finding the optimal number $k$ of species per site to predict.

\subsection{MaxExp framework}
Given \textbf{A1}, the similarity function can be reformulated as depending on the different set values:
\begin{equation}\label{eq:Uprime}
U(y,y') = U'\big( |y\cap y'|,\ |y'|-|y\cap y'|,\ |y|-|y\cap y'|,\ N-|y\cup y'| \big)
= U'(\mathrm{TP},\mathrm{FP},\mathrm{FN},\mathrm{TN}).
\end{equation}

Moreover, let us denote $\pi_x: \{1, \dots, n\} \rightarrow \{1, \dots, n\}$ a permutation of species index in decreasing order of marginal probability \(\left (\mathbb{P}_{X=x}(Y_i= 1) \right )_{i \leq n}\). Considering an observation $(x, y)$ and $k = |y'|$, the parameter to optimize, let us define the following statistics: 

\begin{align}
S_k &= \sum_{i=1}^k y_{\pi_x(i)},\qquad
S^k = \sum_{i=k+1}^N y_{\pi_x(i)} .
\end{align}

Given \textbf{A2}, $S_k$ and $S^k$ are independent. Since FP and TN can be derived from TP and FN, and since $TP=S_k$ and $FN=S^k$, we can reformulate the optimization problem as finding:

\begin{equation}
k^\star = \arg\max_{k\in\mathbb{N}} \sum_{k_1} \sum_{k_2} \mathbb{P}_{X=x}(S_k =k_1) \mathbb{P}_{X=x}(S^k =k_2) \, U'(k_1, k-k_1, k_2, N-k-k_2)
\end{equation}

Solving this problem  can be achieved in $\mathcal{O}(N^3)$, given that the $2 N^2$ values describing the distribution of $S_k$ and $S^k$ for any $k$ can be computed with a complexity equal or less than $\mathcal{O}(N^3)$. Given \textbf{A2}, consider the following relations :

\begin{equation}\label{eq:chain1}
\mathbb{P}_x (S_{k+1} = k_1) =  \mathbb{P}_x (Y_{k+1} = 1)  \mathbb{P}_x (S_k = k_1 - 1) + (1 -  \mathbb{P}_x (Y_{k+1} = 1))  \mathbb{P}_x (S_k = k_1)
\end{equation}
\begin{equation}\label{eq:chain2}
 \mathbb{P}_x (S^{k-1} = k_2) =  \mathbb{P}_x (Y_k = 1)  \mathbb{P}_x (S^k = k_2 - 1) + (1 -  \mathbb{P}_x (Y_k = 1)) \mathbb{P}_x (S^k = k_2)
\end{equation}

where we denote $\mathbb{P}_{X =x}$  as $\mathbb{P}_x$ for readability reasons. With the probabilities now expressed recursively, computing them does not increase the overall complexity of the maximization procedure. In particular, we adopt the plug-in rule strategy where $\mathbb{P}_x (Y_k)$ is replaced by an estimator of the marginal conditional probability of presence, $\hat \eta: \mathcal{X} \rightarrow [0, 1]^N$. This conditional probability is typically the output of a SDM or a site occupancy model. \\\\

In summary, under assumptions \textbf{A1} and \textbf{A2}, it is possible to maximize the expected similarity score in $\mathcal{O}(N^3)$ given some estimated marginal conditional probabilities. Reduction to quadratic complexity is even possible for some scores, like the $F_1$-score or the Jaccard index with a trick explored for $F_\beta$-measures in (\cite{Nan2012}). \\\\

Also, given a similarity function satisfying \textbf{A1}, we proved that MaxExp yields the best theoretical predictions given some true conditional probability as long as they satisfy \textbf{A2}.  An example of MaxExp implementation can be found in Section A of the Supplementary Material. In the following sections, a comparative analysis is conducted to assess the framework's performance in concrete scenarios compared to different reference methods.

\subsection{Reference binarization methods}

To assess the performance of our MaxExp framework, we performed a comparison against a set of reference binarization methods detailed in Table \ref{ref_methods}. These baselines include both established approaches from the literature on species presence–absence predictions (e.g., methods (1), (2), and (3) \cite{Nenzen2011}) and conceptually distinct strategies that we constructed for this study, such as thresholding presence-only predictions (method (4) \cite{Dorm2024}) and from conformal prediction (method (5) \cite{Angelopoulos2022}). Naturally, our framework does not need any calibration, the only potential use of a validation set would have been for the calibration of the predictive model itself. To assess models' ability to estimate presence probabilities, calibration curves on training and testing data are available in Supplementary Material (Section D). \\
Several studied methods however require to calibrate some parameters, and a calibration set has to be constructed to optimize these parameters according to any score used before predictions. Two distinct experiments have thus been run: original test data have been split in a 0.2/0.8 uniformly distributed, where the first part constitutes a validation set and the second the actual data used for score evaluation. A first test has been made in each study with a calibration of these methods using the train set and the output of the model on the latter. In this case, the calibration only reuses the information from which the model has been trained and does not require additional data. A second one has been carried out, taking this time the validation set for calibration. In the last case, this calibration can be seen as an additional source of information which these supervised methods will benefit from. \ \

\begin{table}[H]
    \centering
    \includegraphics[width=\linewidth]{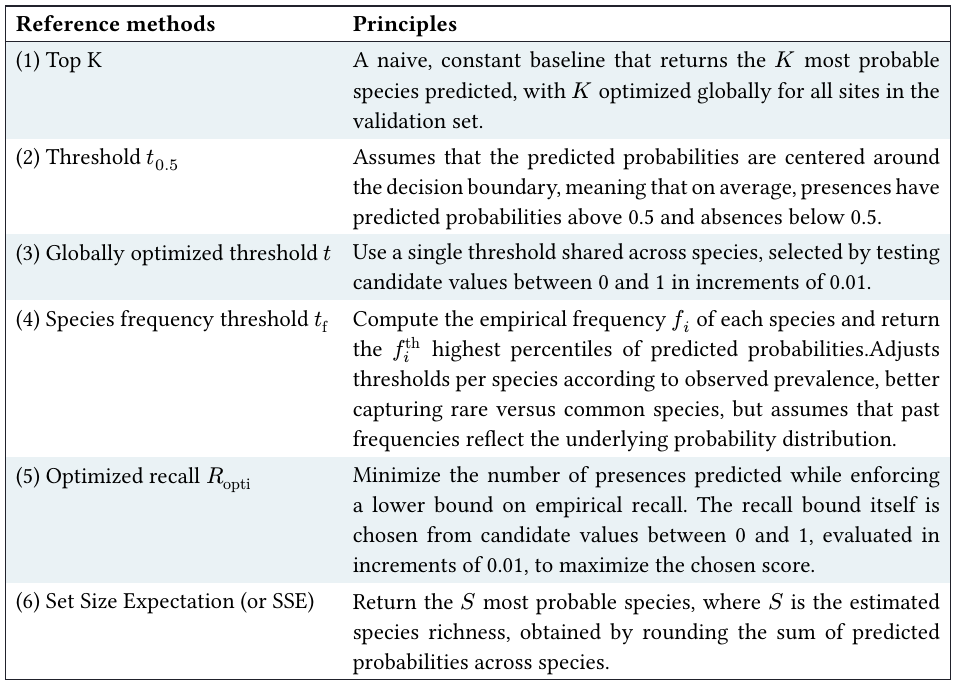}
    \caption{Summary of the different reference methods studied.}
    \label{ref_methods}
\end{table}

In the following, we refer to supervised methods as those requiring the optimization or calibration of one or more parameters. This category includes methods (1), (3), (4) and (5). Methods that do not require parameter optimization or calibration—including our MaxExp framework—are referred to as unsupervised. 

\subsection{Evaluation scores}
As mentioned earlier, our MaxExp framework can be applied to any target score that satisfies assumptions \textbf{A1} and \textbf{A2}. In our experimental analysis, we focused on the following four evaluation scores, commonly used in ecology: : $F_1$, $F_2$, the Jaccard index $J$, and the True Skill Statistic $TSS$.\\

\subsubsection*{$\boldsymbol{F_\beta}$-score ($\boldsymbol{F_\beta}$)}
\begin{equation}
 F_{\beta}=\dfrac{(1+\beta^{2})\mathrm{TP}}{(1+\beta^{2})\mathrm{TP}+\beta^{2}\mathrm{FN}+\mathrm{FP}}
\end{equation} \\

The $F_{\beta}$-score defines a parametrized score, with $\beta$ adjusting the weight of the recall and so the weight of false negatives. The $F_{1}$-score, corresponding to the unweighted mean, is widely used in ecology (\cite{Legendre2012} \cite{Li2013} \cite{Hellegers2025}). The $F_2$-score, more rarely employed, accentuates the weight of false negatives and consequently favors predictions with an increased number of positives predicted, as seen in \Cref{metrics_prev}. In ecological applications where false negatives is more costly than false negatives, this score reflects well the attended behavior for the predictions.

\subsubsection*{Jaccard Index ($\boldsymbol{J}$)}
\begin{equation}
J=\dfrac{\mathrm{TP}}{\mathrm{TP}+\mathrm{FN}+\mathrm{FP}}
\end{equation}\\
The Jaccard index is commonly used in ecological studies — from community comparisons (\cite{Legendre2012}) to fragmentation analyses (\cite{Laurance2007}) — and in classification tasks.
\subsubsection*{True Skill Statistic ($\boldsymbol{TSS}$)}
\begin{equation}
\mathrm{TSS}=\dfrac{\mathrm{TP}}{\mathrm{TP}+\mathrm{FN}}+\dfrac{\mathrm{TN}}{\mathrm{TN}+\mathrm{FP}}-1
\end{equation}\\
Popularized by \cite{Allouche2006} for its perceived independence from species prevalence (later refuted by \cite{Leroy2018})), the TSS is still widely used in ecological modeling. Contrary to $F_{\beta}$ and $J$, $TSS$ evaluates TN and is consequently sensitive to the double-zero problem (\cite{Legendre2012}), which limits its ecological interpretability and explains its particular behavior with unbalanced classes (\Cref{metrics_prev}).

\section{Applications}
The analysis of our MaxExp framework is conducted for three case studies, each evaluated across multiple decision functions and performance metrics. These cases studies were selected to maximize the diversity of testing conditions, thereby assessing both the robustness and flexibility of the proposed framework. The three case studies span distinct taxonomic groups, employ different modeling techniques, and vary by several orders of magnitude in both the number of species and sites. Each study builds upon previous work (\cite{Picek2024}, \cite{Morand2025} and  \cite{Zipkin2023a}), and only the key methodological aspects are presented here. For further details on the data and modeling processes, please refer to the respective original studies. 

\subsection{Case study 1: GeoPlant 2024}
The first case study is derived from the GeoLifeCLEF 2024 challenge, a yearly machine learning challenge focused on the European plant species from the GeoPlant 2024 dataset \cite{Picek2024}. The training data comprises presence-absence (PA) surveys, totaling approximately 90,000 samples and covering 5,016 species of the European flora for presence-absence observations. 
The input predictors used for species prediction fall into three main categories:
\begin{itemize}
    \item Satellite image patches: 3-band (RGB) and 1-band (NIR) images with 128×128 pixel resolution at 10 meters.
    \item Satellite time series: Up to 20 years of temporal Landsat data for six bands (R, G, B, IR, SWIR1, SWIR2), encoded as datacubes with format (years×months×bands).
    \item Climatic Data Cubes: Climatic variables from the CHELSA dataset, more precisely monthly precipitation and min, max, and mean temperature aggregated as cubes of shape (years×months×variables).
\end{itemize}

The training and testing data splits follow the original challenge protocol, employing a spatial block hold-out with 10×10 km grid cells. The model used corresponds to the multimodal baseline provided for the challenge and is available via \href{https://www.kaggle.com/code/picekl/sentinel-landsat-bioclim-baseline-0-31626}{this link}. It consists of an ensemble of two ResNet-18-based encoders for Landsat and climatic data cubes, combined with a Swin-v2-t transformer model for satellite imagery. For the scenario involving calibration on the training set, a 10\% subsample of the training data is used to reduce computational and memory overhead during calibration.

\subsection{Case study 2: Reef Life Survey}

The second case study investigates an other DeepSDM model this time applied to the Reef Life Survey dataset on coral reef fishes (\cite{Edgar2014}). The dataset is composed of 27.437 underwater surveys of coastal habitats in Australia. Observations are then split randomly with a 60/20/20 percentage repartition for train/validation/test datasets, making sure that all surveys from a given location (as defined in the Reef Life Survey metadata) are in the same fold to limit spatial overlap.
The original data being species abundances, it was binarized in the referenced study, resulting in a 1796-species presence-absence dataset. 

The model is composed of 2 heads using respectively a time series of temperature anomaly (Degree Heating Week), and 19 rasters of instantaneous environmental conditions, 15 depicting the natural seascape, and 4 assessing the impact of human activities. Both heads rely on a ResNet-18 encoder. This case study is based on the principle described in \href{https://doi.org/10.24072/pcjournal.471}{this article}, with specifics of the model described (\cite{Morand2025}).

\subsection{Case study 3: American birds}
The third case study builds upon the first scenario of the referred paper (\cite{Zipkin2023a}), which aims to spatially extrapolate species distributions for 27 North American bird species using data from two major citizen science programs: the Breeding Bird Survey (BBS) (\cite{Hudson2017}) and eBird (\cite{Sullivan2014}).

The combined dataset includes 10,383 eBird checklists (used as presence only data) and 356 BBS routes that were binarized as presence-absence records. All observations were collected in the northeastern United States during the first three weeks of June. The data were aggregated into 5×5 km grid cells, resulting in 356 spatial sites. Contrary to the original study, for the train/test split, a 10-fold spatial block cross-validation was applied to the BBS data. One fold was retained for validation and another for testing, while all eBird data were included in the training set (see Section B in Supplementary Material for the maps). Predictions from all trained models on the test fold were aggregated and evaluated against the BBS ground truth. For calibration on train, the mean probability across models trained on the respective fold was used.

\section{Results}

The results for each case study are summarized in (\Cref{score1}, \Cref{score2}, \Cref{score3}). For supervised methods, the parameters of each method were optimized for each column in the table, specifically to maximize the corresponding score. 
As mentionned before, we considered two calibration scenarios: using the training set for calibration, or using a separate validation set. The first scenario allows a more direct and fair comparison with unsupervised methods, as it does not provide additional information beyond the training data.

For each test conducted, a permutation test was performed to assess whether the mean score achieved by the proposed framework is significantly higher than that of the other methods (see Section C in the Supplementary Material).

\subsection{Case Study 1}
MaxExp achieves significantly higher scores than all alternatives across all scores (permutation test, p < 0.05), except SSE on $F_1$ and Jaccard, where differences are not statistically significant. Only when calibrated on the validation set, the global threshold method Th.\,$t$  becomes competitive. When calibrated on training set, its performance degrades significantly.
\begin{table}[H]
    \centering
    \includegraphics[width=0.9\linewidth]{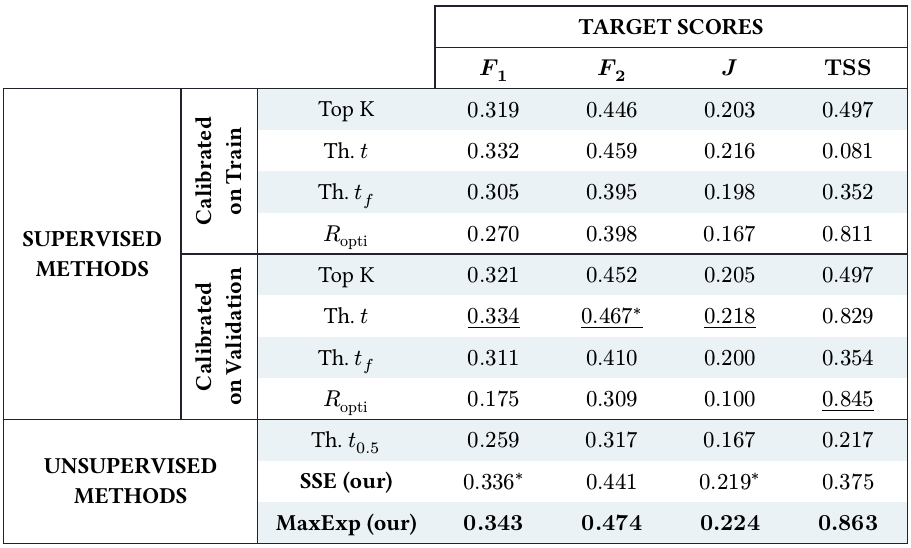}
\caption{Mean scores obtained for the different methods (\Cref{ref_methods}) applied to Case Study 1. 
Bold indicates the best unsupervised scores. Underlined values indicate best supervised 
scores. Asterisks mark results within the confidence interval of MaxExp (p < 0.05).}
\label{score1}
\end{table}

\subsection{Case Study 2}
In this experiment, calibration on the validation dataset substantially improved the scores of supervised methods. While MaxExp remains the top-performing unsupervised method (and also outperforms supervised methods calibrated on the training set), the globally optimized threshold method calibrated on the validation set achieves higher scores across all metrics except TSS.

\begin{table}[H]
    \centering
    \includegraphics[width=0.9\linewidth]{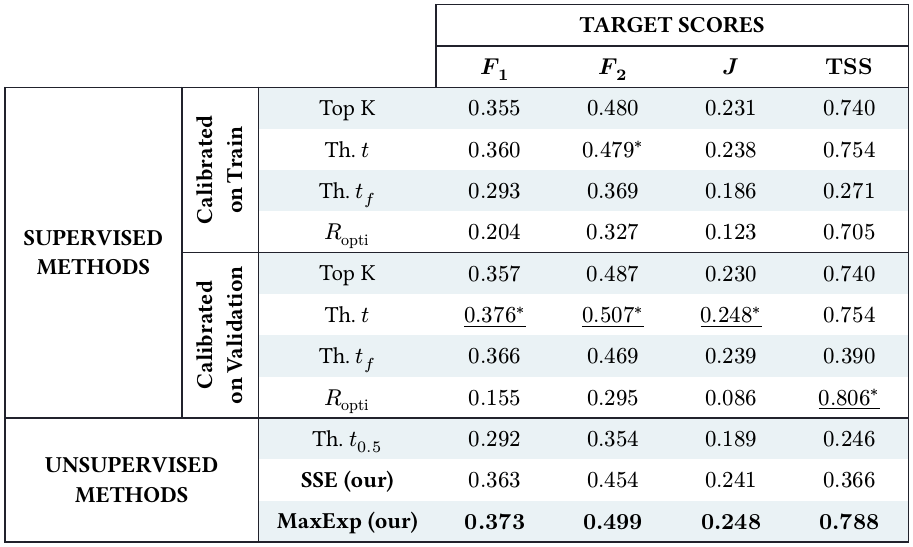}
\caption{Mean scores obtained for the different methods applied to Case Study 2. 
Bold indicates the best unsupervised scores. Underlined values indicate best supervised 
scores. Asterisks mark results within the confidence interval of MaxExp (p < 0.05).}
\label{score2}
\end{table}

\subsection{Case Study 3}
In this third case study, MaxExp globally outperforms all other methods; only the globally optimized threshold method—whether calibrated on the training or validation set—achieves competitive performance.

\begin{table}[H]
    \centering
    \includegraphics[width=0.9\linewidth]{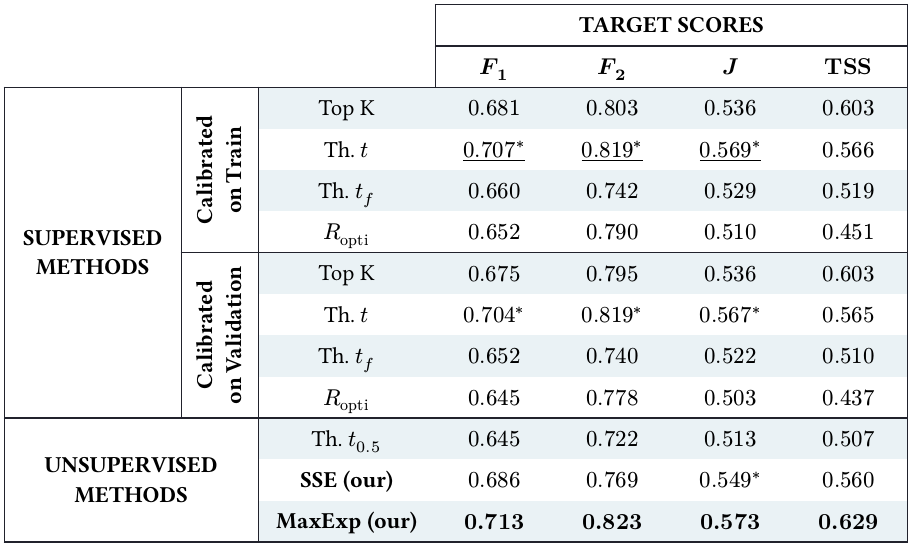}
\caption{Mean scores obtained for the different methods applied to Case Study 3. 
Bold indicates the best unsupervised scores. Underlined values indicate best supervised 
scores. Asterisks mark results within the confidence interval of MaxExp (p < 0.05).}
\label{score3}
\end{table}

Across all three case studies, the MaxExp framework consistently improves performance across all evaluated scores among unsupervised methods and those calibrated on train set. Notably, only the test-based supervised global threshold method outperformed MaxExp on case study 2, will the latter achieve statistically significant but inferior scores in most scenarios when compared to MaxExp, using either test and train dataset for calibration.\\

The Set Size Expectation (SSE) method also achieves substantial performance for both the $F_1$-score and the Jaccard index ($J$) across all of the 3 case studies. This illustrates how the number of species predicted per site—based on the sum of predicted probabilities interacts with these metrics, even though the estimated richness can be biased and highly variable in practice.

\section{Discussion}

\subsection{Result analysis}
Our study demonstrates that SDM binarization — the critical step from species probability of occurrence in a given site to presence–absence prediction — can be addressed as a decision optimization problem rather than as a heuristic approach. By adopting this change of paradigm, through the introduction of the new MaxExp framework, we show that unsupervised multispecies predictions can achieve better performances than classical methods across diverse taxa and ecosystems without additional calibration. A second insight is that the simple Set Size Expectation (SSE) baseline is surprisingly effective, highlighting that interpretability and computational efficiency are not necessarily at the expense of predictive accuracy. Together, these findings provide ecologists with complementary tools for multispecies predictions: MaxExp as a general-purpose optimizer adapted to a large range of scores, and SSE as a lightweight, intuitive alternative for base target scores such as F1 and Jaccard. \\

Our three case studies also highlight that dataset characteristics significantly condition the relative performance of binarization methods. First, for the GeoPlant dataset, which is species-rich (5,000+ taxa), spatially heterogeneous, and strongly imbalanced with a long tail of rare species (\cite{Picek2024}), MaxExp clearly outperforms other methods across all scores, highlighting its relevance for high-dimensional and imbalanced conditions, which are common in ecology. SSE also performs well for $F_1$ and Jaccard, reflecting the close alignment of predicted and observed richness with these scores in highly diverse communities. \\ 

In contrast, for the Reef Life Survey fish dataset (\cite{Edgar2014}), which presents a smaller number of sites but still a high number of species. the supervised thresholding methods calibrated on a validation set perform better than on GeoPlant, in some cases rivaling with MaxExp. This suggests that in datasets with strong distribution shifts between train and test datasets, the use of supervised calibration similar to the test sets can partially compensate for miscalibrated probability estimates. \\
Alternatively, one could have calibrated the predictive models themselves on the validation set and then applied MaxExp to the resulting probabilities. This would likely have yielded higher performance than the supervised methods, as these methods already achieve results comparable to MaxExp in its unsupervised form. \\

Finally, for the North American bird dataset including relatively few species and sites, with data aggregated from standardized monitoring (BBS) (\cite{Hudson2017}) and citizen science (eBird) (\cite{Sullivan2014}), MaxExp outperforms reference methods for every scores. The performance gap between thresholding approaches is nonetheless smaller than in the previous case studies. We hypothesize that the limited number of species, restricted geographical region, and narrow sampling timeframe produced a dataset with low variance, reducing the advantage of unsupervised methods. This hypothesis is supported by a table of scores higher and far more homogenous than for previous cases, and by a model with the best calibration curve on test of the three cases, despite being the simplest (Figure 4, Supplementary Material). The smaller taxonomic and spatial scope, combined with better-calibrated probabilities results in the global threshold method performing competitively, and differences between MaxExp, SSE, and calibration-based methods were less pronounced.

\subsection{Metric choice}

Another important consideration, as emphasized in the introduction (see \Cref{metrics_prev}), is that the selection of a specific evaluation metric inherently determines the nature of the resulting predictions. The optimization process, either unsupervised or related to a calibration step, is metric-dependent, and different target scores can lead to substantially divergent spatial and statistical patterns of predicted species prevalence (\cite{Hellegers2025}) or biodiversity composition. Each evaluation metric embodies a distinct conceptualization of predictive accuracy—prioritizing, for example, the detection of rare species, the minimization of false positives, or the achievement of realistic overall prevalence estimates.\\

Within this framework, the $F_\beta$ family of scores provides a clear illustration of how varying the parameter $\beta$ allows one to adjust the relative importance of false negatives and false positives. Although the functional relationship between $\beta$ and its influence on the optimized prediction thresholds is not analytically straightforward, empirical evidence indicate that higher values of $\beta$ tend to produce larger predicted species sets (\Cref{metrics_prev}). This reflects an increased tolerance for false positives while reducing the number of false negatives. \\

While the balanced $F_1$-score remains the predominant choice in the literature, adopting higher $\beta$ values may be particularly relevant in ecological contexts where false negatives are more detrimental than false positives. Such situations include, for example, the assessment of potential favorable habitat or refuges for rare or endangered species (\cite{Pichot2025}, \cite{Semper-Pascual2023}, or scenarios where repeated non-detections compromise confidence in absence predictions.

\subsection{Limitations and Perspectives}
The present study focused on sample-averaged metrics, i.e., metrics that measure model performance over species assemblages per site. However, MaxExp can also be applied to macro-averaged metrics, i.e., metrics that measure model performance over mean prevalence per species, then averaged across species. This extension is partially exposed in Section E of the Supplementary Material. However, if theoretical results are invariant to the transposition of the problem, our tests conclude that the robustness to concrete cases of the MaxExp framework may not be translated to the macro-averaged score optimization. If MaxExp still leads to performance improvement in several experiments, this difference is not anymore systematic and the results vary from case studies and from chosen scores. The scores being significantly lower than their sample-average equivalent, the error in probability estimates for some species impact more greatly macro-average scores. Besides, if Top K method for macro-averaged scores is not adapted due to strong species prevalence, species-related threshold method Th. $t_s$  can in this case be computed, and can seemingly lead to improved performances for validation-based calibration. \\

A second limitation of MaxExp lies in its implicit assumption of species independence, which overlooks ecological interactions such as facilitation or competition (\cite{Steen2014}). In this respect, MaxExp aligns with most binarization methods found in the literature. However, recent studies have sought to move beyond this assumption, for instance with the threshold classifier proposed by (\cite{Dorm2024}). Addressing these challenges — e.g., by integrating species co-occurrence models, trait-based approaches, or new evaluation scores — represents an important direction for future research. \\

Regarding methodological perspectives, future work could explore the comparison with end-to-end learning of score-aware predictors (\cite{Calabrese2014}), where the optimization of evaluation criteria is embedded directly into the training phase of the models rather than in a post hoc binarization. 
Finally, the explicit link between decision rules and evaluation scores invites a broader assessment of how the choice of scores aligns with predictive tasks, highlighting the importance of a thoughtful evaluation metric selection and opening the way to score design tailored to conservation priorities, notably emphasizing endangered or rare species (\cite{Mouillot2024}).\\

In conclusion, two main messages emerge. First, MaxExp provides a flexible, unsupervised optimization framework that improves presence–absence predictions for most commonly used evaluation scores of species assemblage predictions. SSE additionally offers a simple and efficient alternative which, while not always matching MaxExp, performs competitively under two of the four scores considered. Together, these approaches reframe binarization not as a heuristic compromise but as a transparent and reproducible decision process that can be explicitly aligned with ecological objectives of decision-making (\cite{Guisan2013}, \cite{Kass2024}).

\section{Acknowledgements}
The research described in this paper was funded by the European Commission through the MAMBO project (grant agreement 101060639) and the GUARDEN project (grant agreement 101060693). The authors would also want to thanks the CNRS GDR OMER for its support and resources.

\section{Conflict of Interest}
The authors declare no conflicts of interest.

\printbibliography

\end{document}


\linenumbers
\section{Supplementary Material}
\appendix
\section{MaxExp Algorithm}\label{Algo}

The MaxExp algorithm.  No assumption about the score $U$ is presumed besides \textbf{A1} and \textbf{A2}. Consequently the resolution is in $\mathcal{O}(N^3)$ complexity, where $N$ denotes the number of species of interest. It takes in input the estimated probability vector $\hat \eta$ and the score $U$ and return the set of predicted species. Assumption \textbf{A1} and \textbf{A2} allows us to stop at the first expected score maximum, diminishing greatly in average the number of expected score $S(K)$ to compute.

\begin{algorithm}[H]
\caption{MaxExp($\hat{\eta}$, $U$)}
\SetAlgoLined
\DontPrintSemicolon
\KwIn{$\hat{\eta}$, $U$}
\KwOut{$K^{\text{th}}$ most probable species}
\;
1. Compute the probability matrices\;
\;
$P, P' \gets O_{N+1}, O_{N+1}$\;
$\eta \gets \text{sort}(\hat{\eta}, \text{'decreasing'})$\;
$P_{1,1}, P'_{1,1} \gets 1, 1$\;
\For{$i \in [\![1,N]\!]$}{
    $P_{i+1,0} \gets (1-\eta_i)P_{i,0}$\;
    \For{$j \in [\![1,i+1]\!]$}{
        $P_{i+1,j+1} \gets \eta_i P_{i,j} + (1-\eta_i)P_{i,j+1}$\;
    }
}
\For{$i \in [\![1,N]\!]$}{
    $P'_{i+1,0} \gets (1-\eta_{N-i+1})P'_{i,0}$\;
    \For{$j \in [\![1,i+1]\!]$}{
        $P'_{i+1,j+1} \gets \eta_{N-i+1} P'_{i,j} + (1-\eta_{N-i+1}) P'_{i,j+1}$\;
    }
}
\;
2. Compute expected score $S(K)$ with $K$ running from 1 to first maximum\;
\;
$S^*, S, K^*, K \gets -\infty, -\infty, 0, 0$\;
\While{$K \leq N$ and $S = S^*$}{
    $S \gets 0$\;
    \For{$i \in [\![1,K+1]\!]$}{
        \For{$j \in [\![1,N-K+1]\!]$}{
            Increment $S(K)$ along $U(\text{TP},\text{FP},\text{FN},\text{TN})$\;
            $S \gets S + P_{K,i} S_{N-K,j} \times U(i, K-i, j, N-K-j)$\;
        }
    }
    \If{$S > S^*$}{
        $S^*, K^* \gets S, K$\;
    }
}
\Return{$K^{\text{th}}$ most probable species}
\end{algorithm}

\section{Maps American Birds}
\begin{figure}[H]
    \centering
    \begin{subfigure}[t]{0.45\textwidth}
        \centering
        \includegraphics[width=\linewidth]{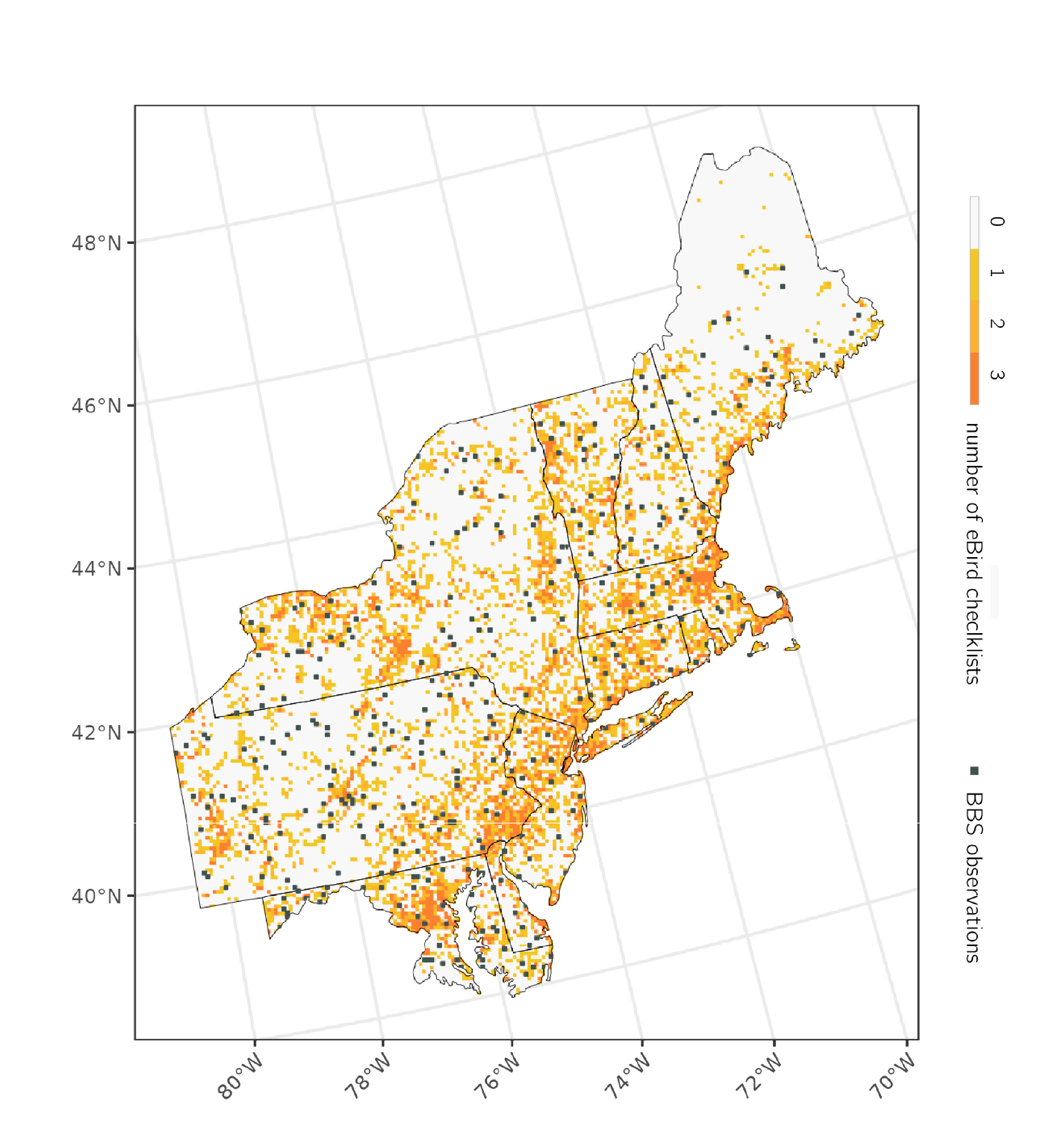}
        \caption{Density of Bird checklists and location of the BBS routes.}
        \label{fig:data_loc}
    \end{subfigure}
    \hspace{0.5cm}
    \begin{subfigure}[t]{0.45\textwidth}
        \centering
        \includegraphics[width=\linewidth]{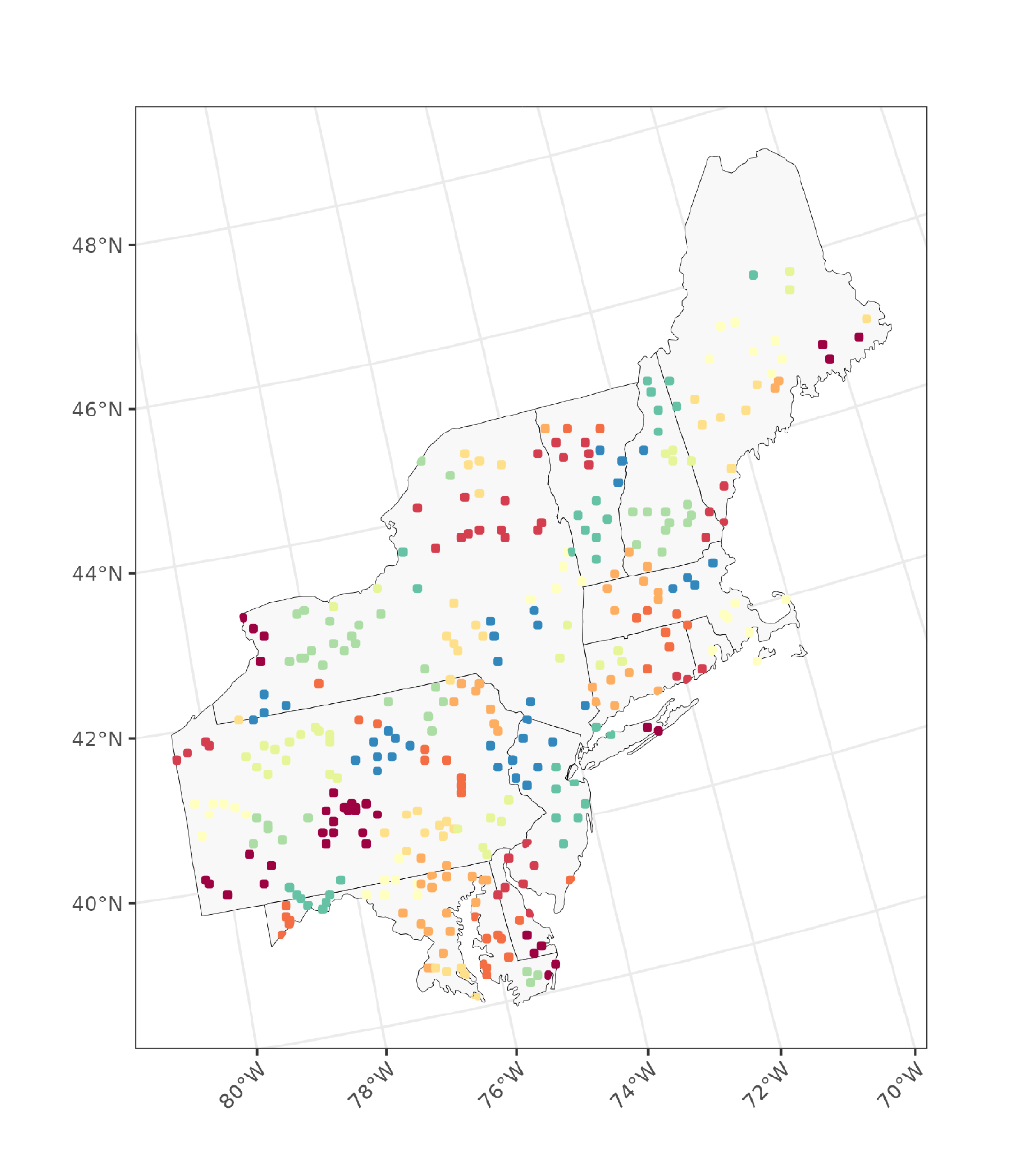}
        \caption{Folds from BBS checklists, determined by a group of dots (representing sites) of the same color}
        \label{fig:folds_loc}
    \end{subfigure}
    \caption{Distribution of sites and observations used in the third case study.}
    \label{birds_data}
\end{figure}

\section{Permutation test p-values}

\begin{table}[H]
    \centering
    \includegraphics[width=0.9\linewidth]{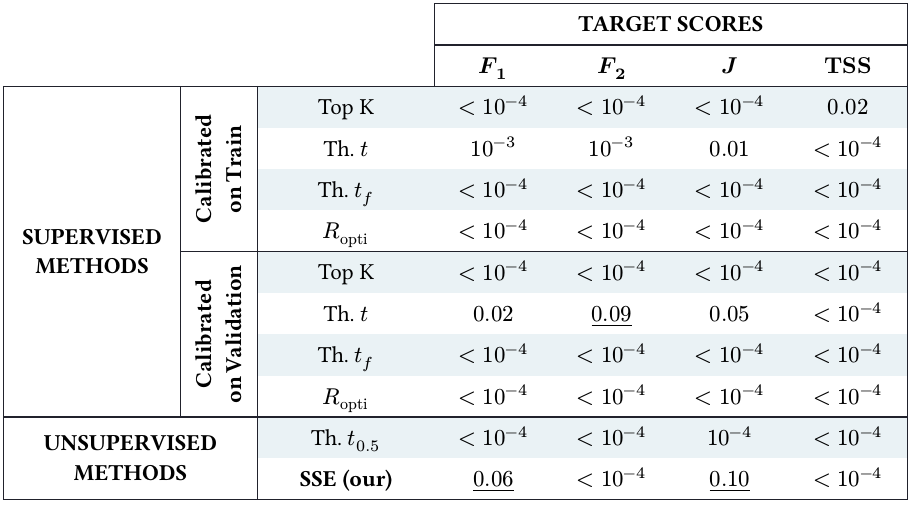}
\caption{p-values from permutation test of scores $S$ ($H_1 : \mathbb{E}[S_{\text{MaxExp}}] > \mathbb{E}[S]$, $n_{\text{per}} = 9999$). For p-values $< 0.01$, only order of magnitude is shown. Cases where the score is statistically significant are underlined.}
\label{pvalue1}
\end{table}

\begin{table}[H]
    \centering
    \includegraphics[width=0.9\linewidth]{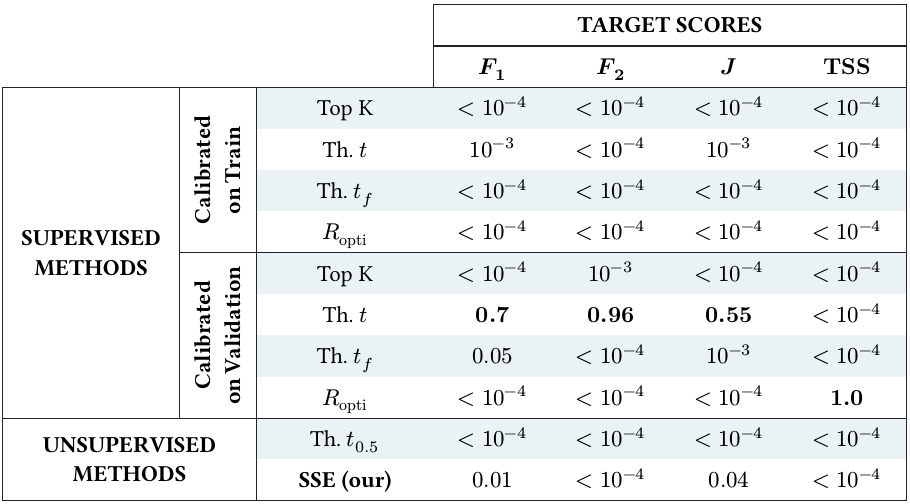}
\caption{p-values from permutation test for Case Study 2. Format and significance markers as in Table \ref{pvalue1}.}
\label{pvalue2}
\end{table}

\begin{table}[H]
    \centering
    \includegraphics[width=0.9\linewidth]{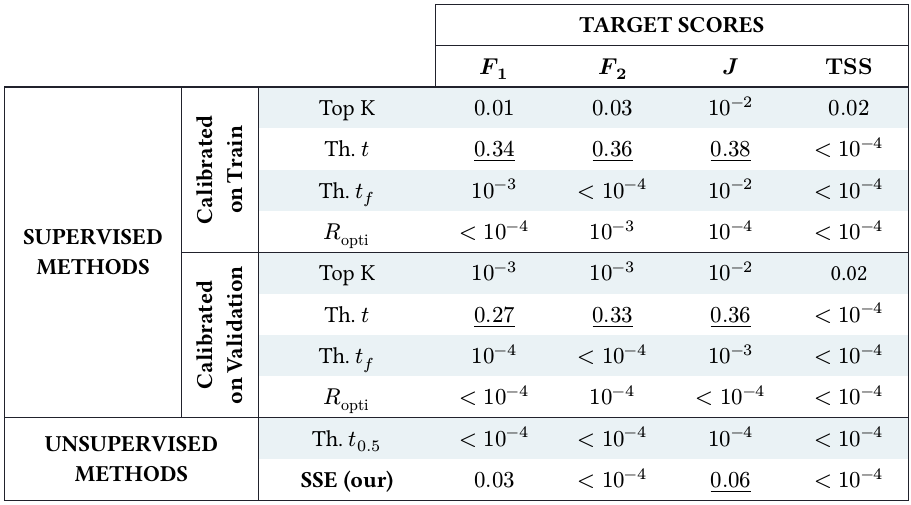}
\caption{p-values from permutation test for Case Study 3. Format and significance markers as in Table \ref{pvalue1}.}
\label{pvalue3}
\end{table}

\section{Calibration curves}\label{Calib}

\begin{figure}[H]
    \centering
    \begin{subfigure}[t]{0.45\textwidth}
        \centering
        \includegraphics[width=\linewidth]{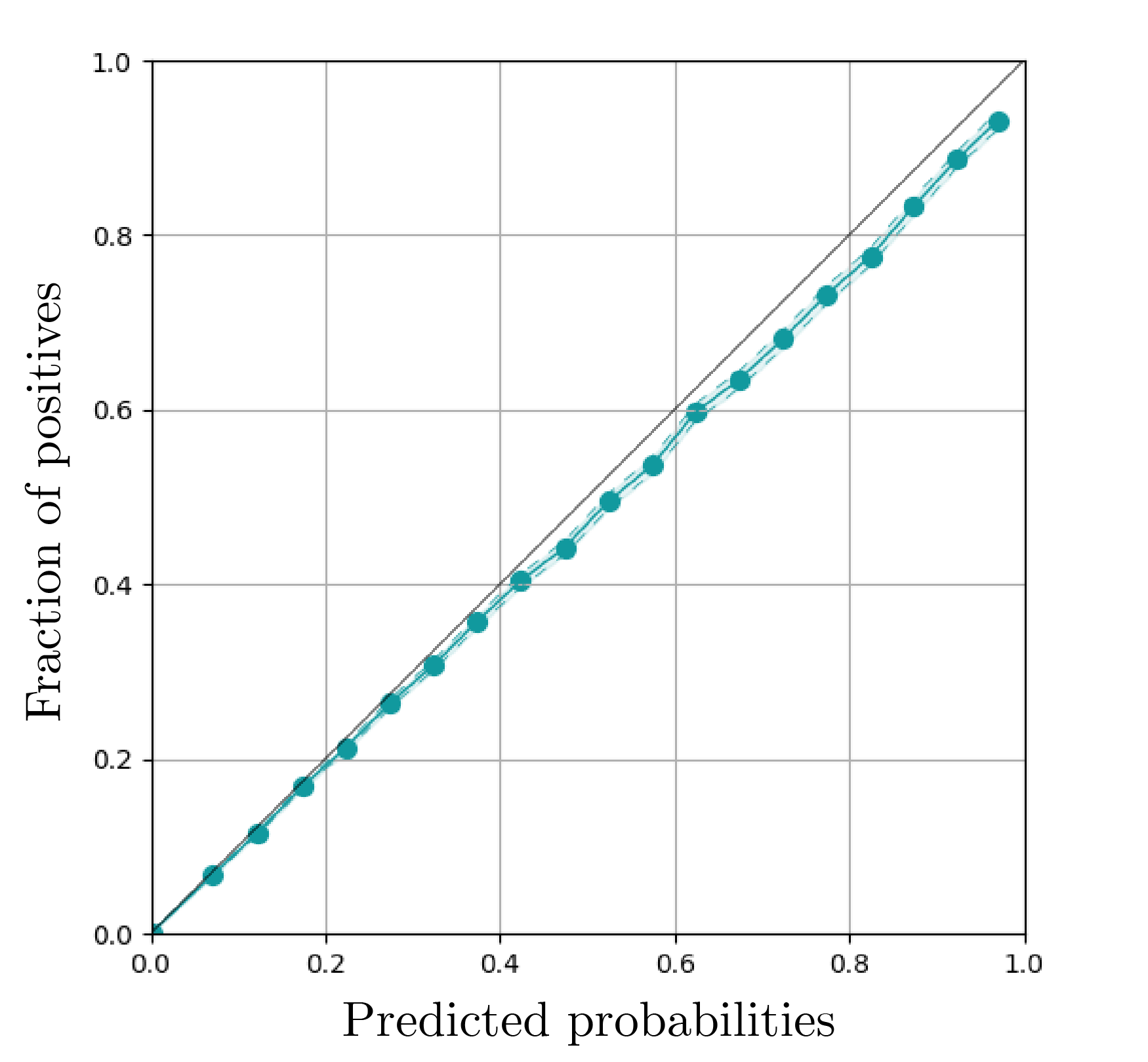}
        \caption{Calibration curve from train data}
        \label{calib_train_1}
    \end{subfigure}
    \hspace{0.5cm}
    \begin{subfigure}[t]{0.45\textwidth}
        \centering
        \includegraphics[width=\linewidth]{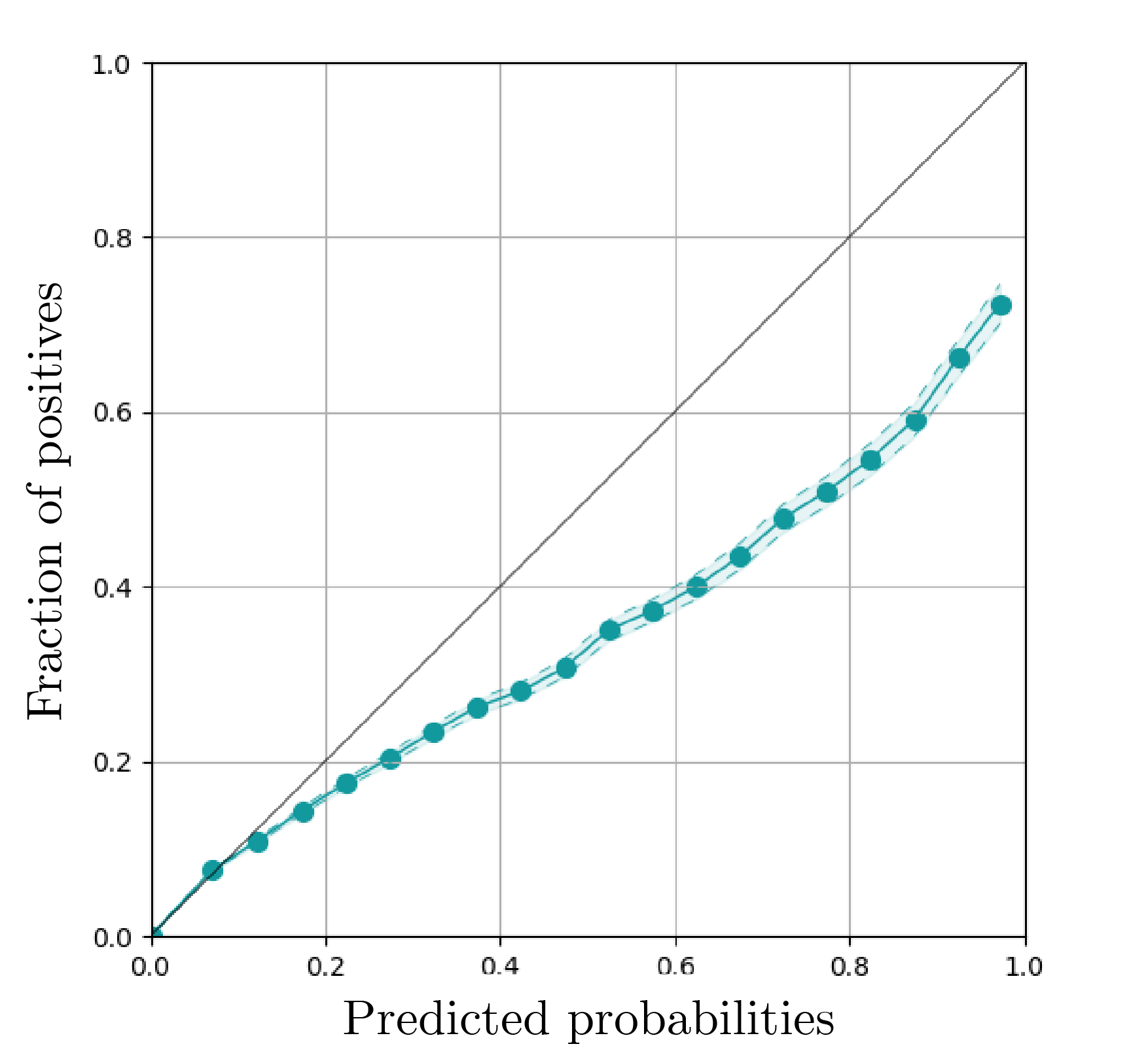}
        \caption{Calibration curve from test data}
        \label{calib_test_1}
    \end{subfigure}
    \caption{Calibration curves from the model and data in the first study. The predicted probabilities across all sites and species are placed in bins, and the fraction of presences are calculated from the related observations. The blue zone corresponds to an estimation of a $\pm 2\sigma$ interval. We can clearly observe over-predictions of the model for the test set, which can affect the viability of MaxExp to find the expected optimum.}
    \label{cc_1}
\end{figure}

\begin{figure}[H]
    \centering
    \begin{subfigure}[t]{0.45\textwidth}
        \centering
        \includegraphics[width=\linewidth]{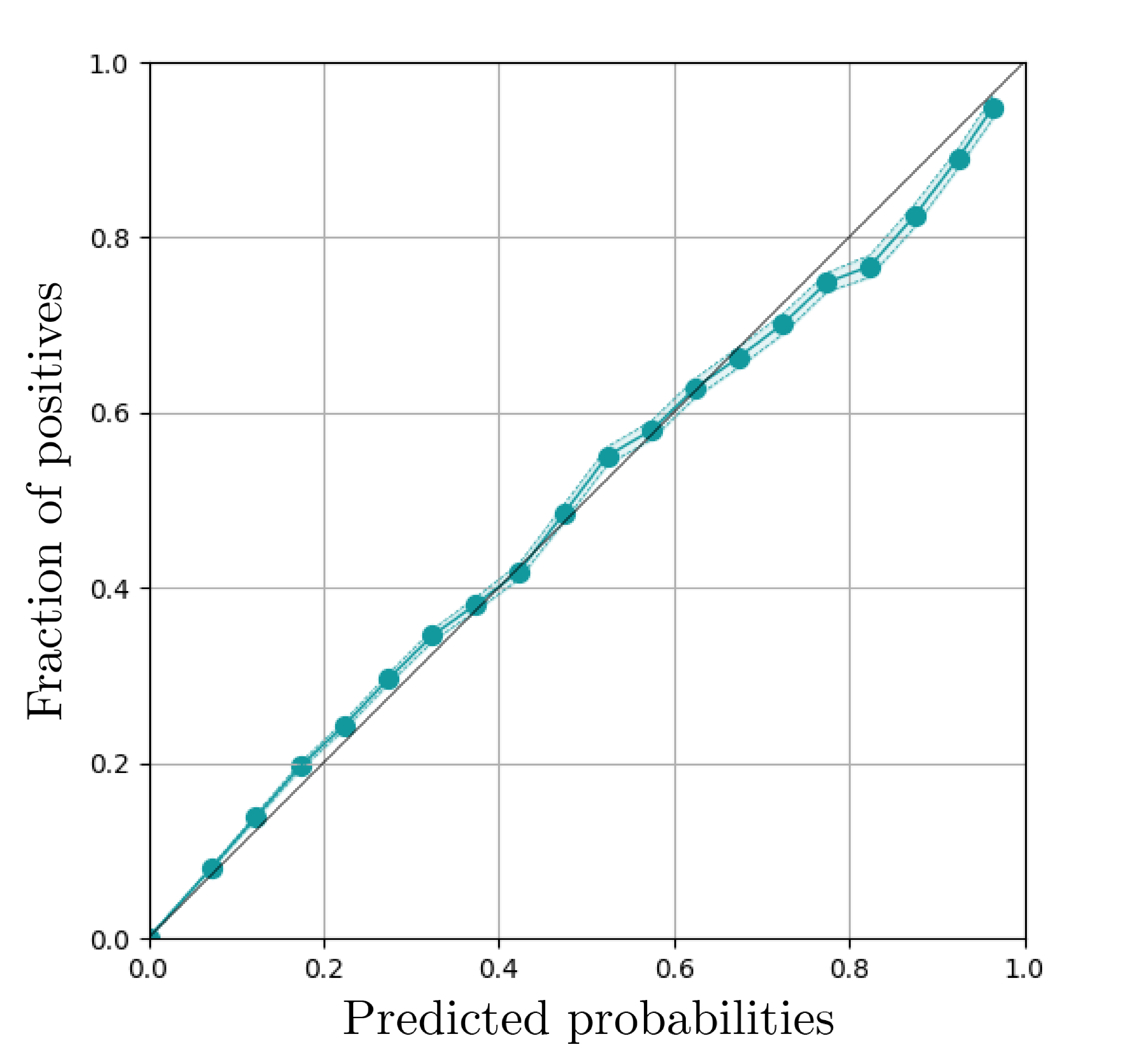}
        \caption{Calibration curve from train data}
        \label{calib_train_2}
    \end{subfigure}
    \hspace{0.5cm}
    \begin{subfigure}[t]{0.45\textwidth}
        \centering
        \includegraphics[width=\linewidth]{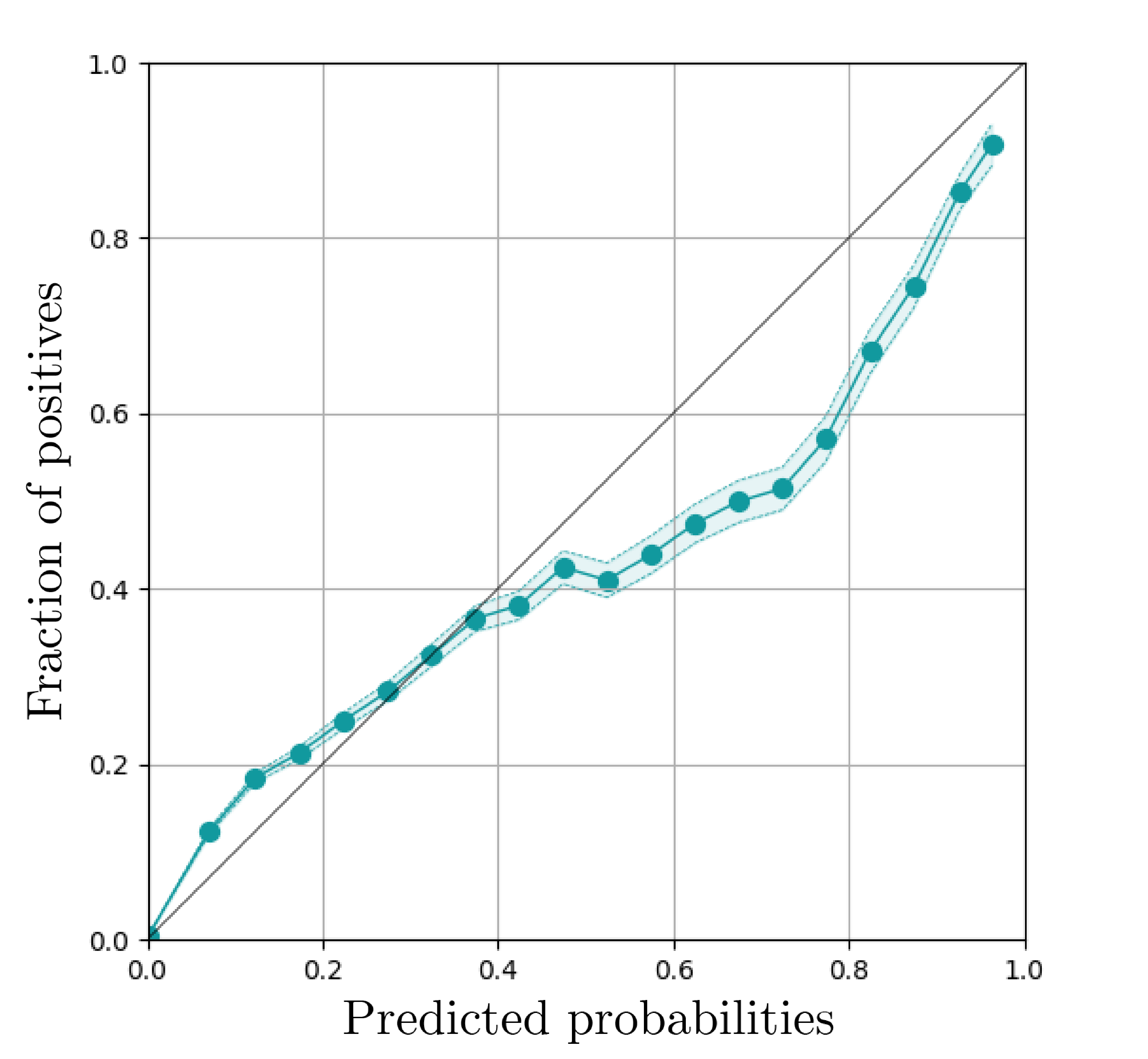}
        \caption{Calibration curve from test data}
        \label{calib_test_2}
    \end{subfigure}
    \caption{Calibration curves from the model and data in the second study. The predicted probabilities across all sites and species are placed in bins, and the fraction of presences are calculated from the related observations. The blue zone corresponds to an estimation of a $\pm 2\sigma$ interval.}
    \label{cc_2}
\end{figure}

\begin{figure}[H]
    \centering
    \begin{subfigure}[t]{0.45\textwidth}
        \centering
        \includegraphics[width=\linewidth]{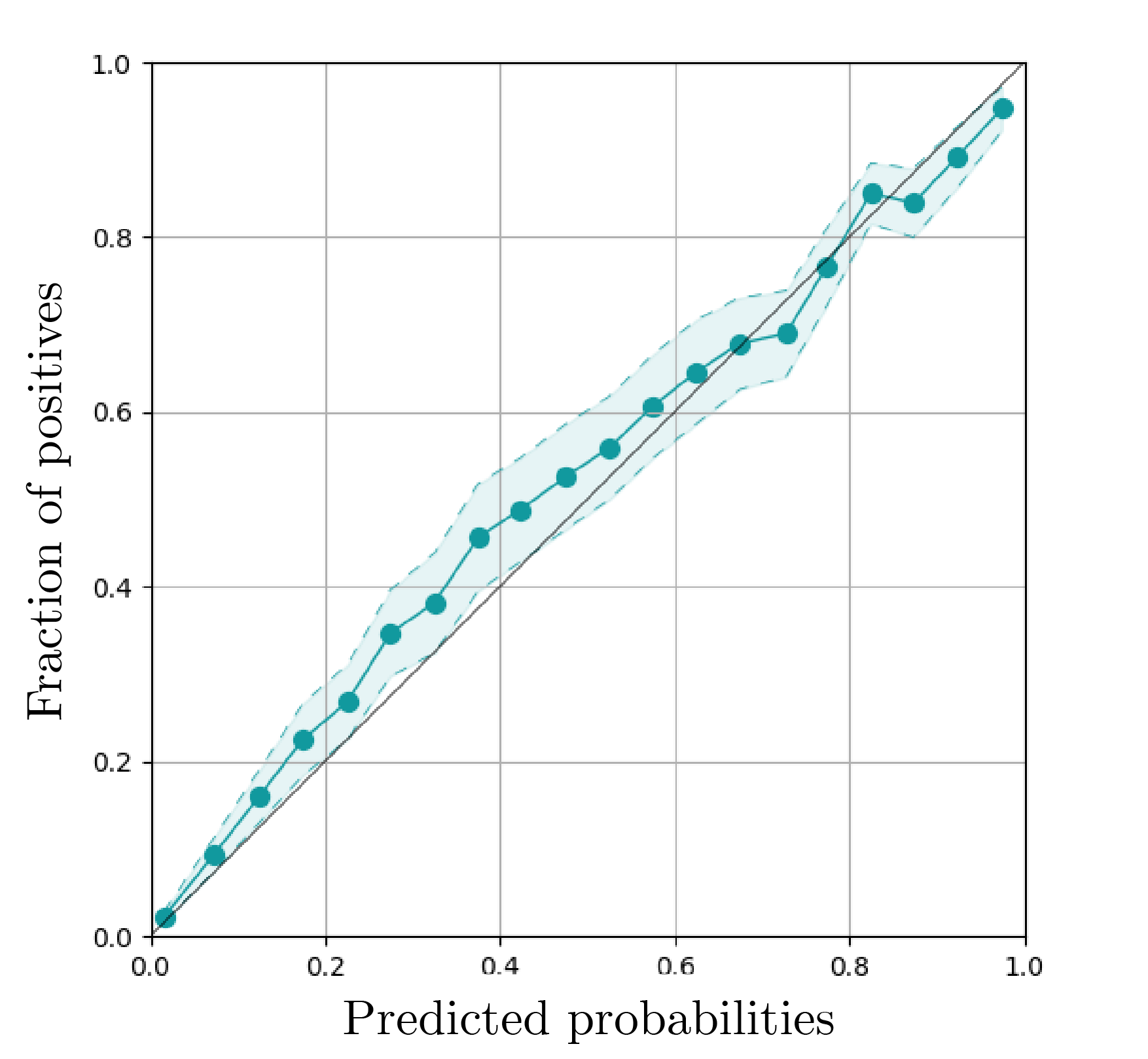}
        \caption{Calibration curve from train data}
        \label{calib_train_3}
    \end{subfigure}
    \hspace{0.5cm}
    \begin{subfigure}[t]{0.45\textwidth}
        \centering
        \includegraphics[width=\linewidth]{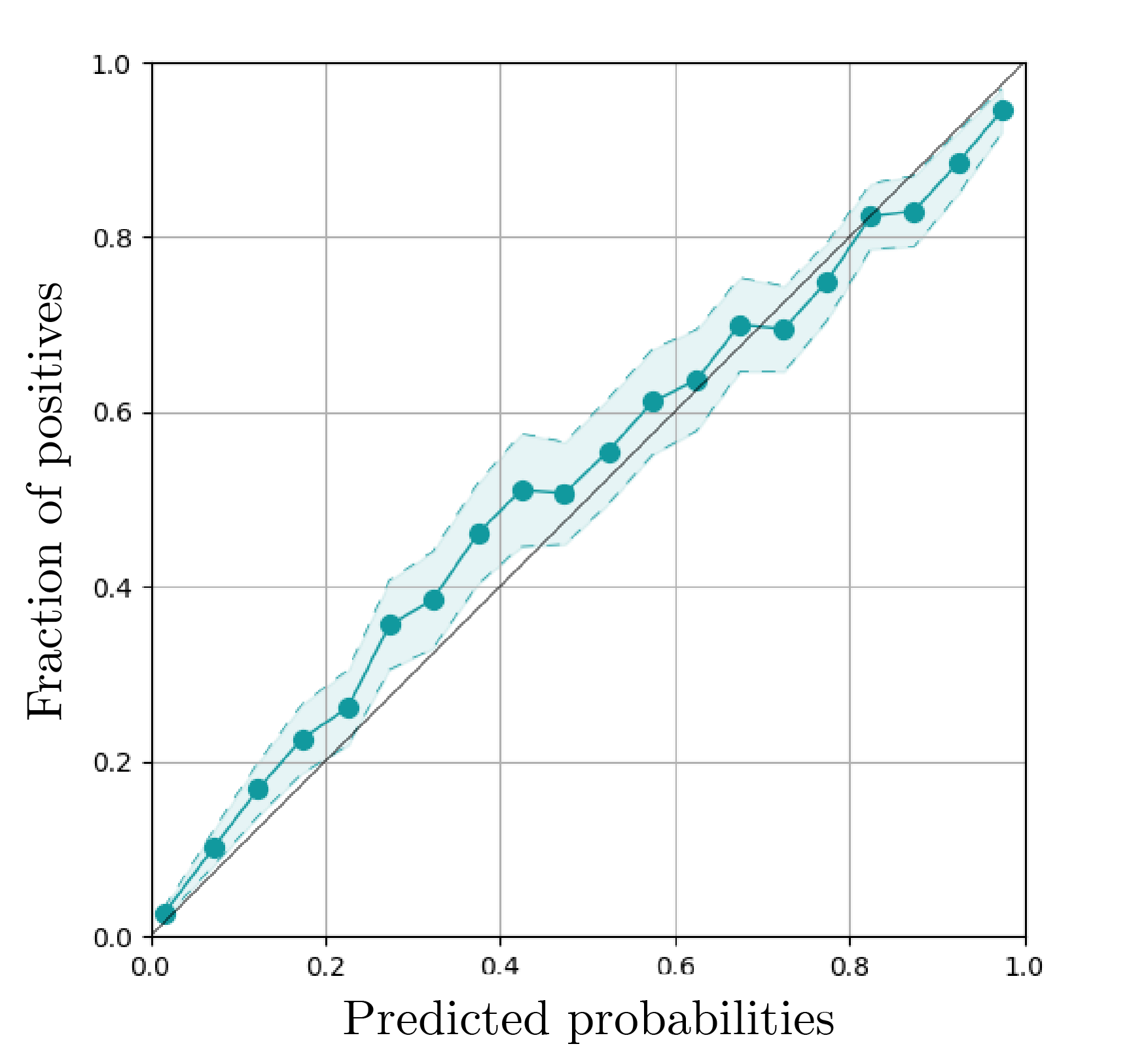}
        \caption{Calibration curve from test data}
        \label{calib_test_3}
    \end{subfigure}
    \caption{Calibration curves from the model and data in the third study. The predicted probabilities across all sites and species are placed in bins, and the fraction of presences are calculated from the related observations. The blue zone corresponds to an estimation of a $\pm 2\sigma$ interval.}
    \label{cc_3}
\end{figure}

\section{Macro-averaged scores}\label{Macro}

\begin{table}[H]
    \centering
    \includegraphics[width=0.9\linewidth]{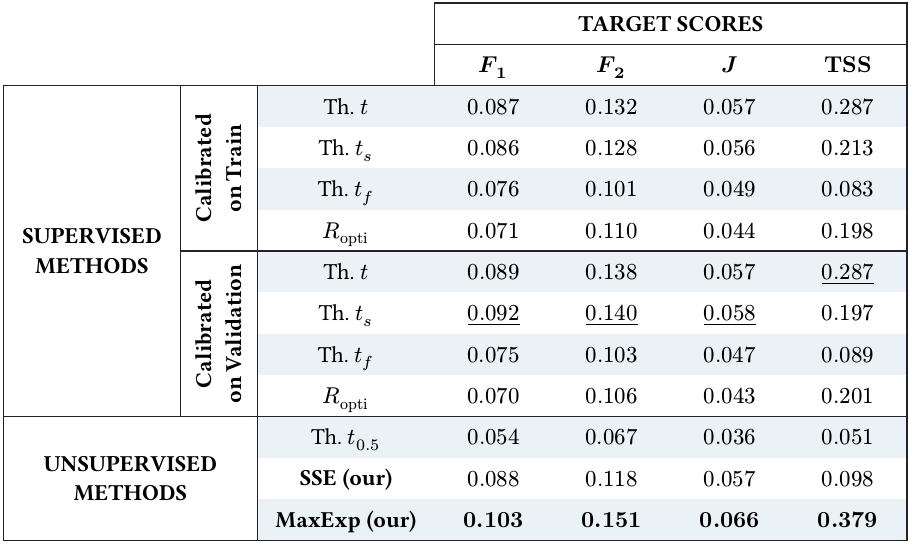}
\caption{Mean of the macro-averaged scores $S_m$ obtained for the different methods applied to Case Study 1. The best scores obtained on unsupervised methods are in bold while those obtained on supervised methods are underlined.}
\label{appendix_d1}
\end{table}

\begin{table}[H]
    \centering
    \includegraphics[width=0.9\linewidth]{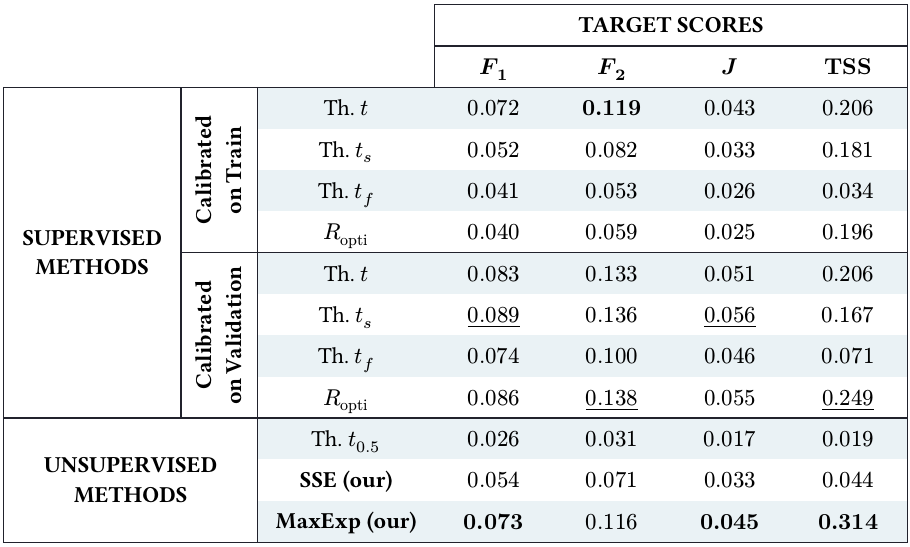}
\caption{Mean of the macro-averaged scores $S_m$ obtained for the different methods applied to Case Study 2. The best scores obtained on unsupervised methods are in bold while those obtained on supervised methods are underlined.}
\label{appendix_d2}
\end{table}

\begin{table}[H]
    \centering
    \includegraphics[width=0.9\linewidth]{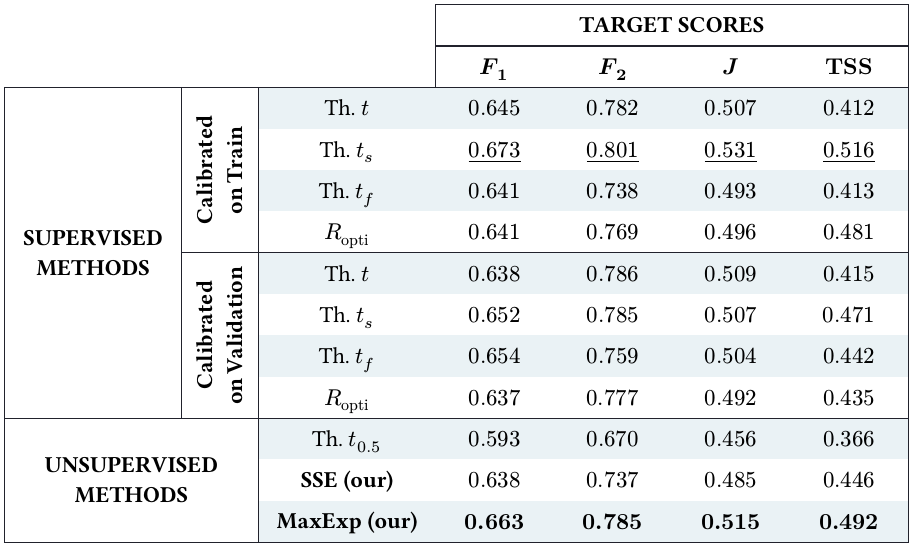}
\caption{Mean of the macro-averaged scores $S_m$ obtained for the different methods applied to Case Study 3. The best scores obtained on unsupervised methods are in bold while those obtained on supervised methods are underlined.}
\label{appendix_d3}
\end{table}